\documentclass{article}
\usepackage{icassp,amsmath,graphicx}
\usepackage{algorithm}
\usepackage{algorithmic}
\usepackage{makecell}
\usepackage{array}
\usepackage{times}  % DO NOT CHANGE THIS
\usepackage{helvet}  % DO NOT CHANGE THIS
\usepackage{courier}  % DO NOT CHANGE THIS
\usepackage[hyphens]{url}  % DO NOT CHANGE THIS
\usepackage{graphicx} % DO NOT CHANGE THIS

\usepackage{color}

\usepackage{flushend}
\usepackage{amsfonts}
\usepackage{amsmath}
\usepackage{amssymb}
\definecolor{dark}{rgb}{0.2, 0.2, 0.5}
\usepackage[pagebackref=false,colorlinks=true,bookmarks=false,linkcolor=red,citecolor=dark,urlcolor=black]{hyperref}
\usepackage{newfloat}
\usepackage{listings}
\let\OLDthebibliography\thebibliography
\renewcommand\thebibliography[1]{
  \OLDthebibliography{#1}
  \setlength{\parskip}{0pt}
  \setlength{\itemsep}{0pt plus 0.3ex}
}

\title{MODIFY: Model-driven Face Stylization without Style Images}
\name{Yuhe Ding$^{1}$, Jian Liang$^{3}$, Jie Cao$^{3}$, Aihua Zheng$^{2}$, and Ran He$^{3}$}
\address{$^{1}$ School of Computer Science, Anhui University
$^{2}$ School of Artificial Intelligence, Anhui University \\
$^{3}$ CRIPAC \& MAIS, Institute of Automation, Chinese Academy of Sciences}
\begin{document}
%\ninept
%
\maketitle
\begin{abstract}
Existing face stylization methods always acquire the presence of the target (style) domain during the translation process, which violates privacy regulations and limits their applicability in real-world systems.
To address this issue, we propose a new method called MODel-drIven Face stYlization (MODIFY), which relies on the generative model to bypass the dependence of the target images.
Briefly, MODIFY first trains a generative model in the target domain and then translates a source input to the target domain via the provided style model.
% In particular, MODIFY employs the encoder-decoder architecture to disentangle the input into content and style.
To preserve the multimodal style information, MODIFY further introduces an additional remapping network, mapping a known continuous distribution into the encoder's embedding space.
During translation in the source domain, MODIFY fine-tunes the encoder module within the target style-persevering model to capture the content of the source input as precisely as possible.
Our method is extremely simple and satisfies versatile training modes for face stylization.
% , \textit{i.e.}, offline, online, and test-time training. 
Experimental results on several different datasets validate the effectiveness of MODIFY for unsupervised face stylization.
Code will be released at https://github.com/YuheD/MODIFY.
% We hope our method will shed light on the future research of image translation.
\end{abstract}
\begin{keywords}
face stylization, test-time training %, transfer learning, image generation
\end{keywords}
\section{Introduction}
Face stylization~\cite{warpgan,psp} aims to translate the face images from the source (content) domain into the target (style) domain.
It has been widely used in popular software such as \textit{Photoshop}, \textit{Instagram}, \textit{Beauty Camera}, and \textit{Tik Tok}.
Recent methods have shown impressive results on unsupervised face stylization~\cite{Men_2022_CVPR,cyclegan,Li_2022_CVPR}.
% \begin{figure}
% 	\begin{center}
% 		\includegraphics[scale=0.2]{results/motivation.pdf} 
% 	\end{center}
% 	\caption{Model-driven face stylization, a new problem setting this work studies. 
% 	The orange line denotes traditional data-driven face stylization methods that require both source and target datasets during training. The purple line denotes the proposed model-driven face stylization, which learns a mapping between the source dataset and the style-preserving model, without the access to target images.}
% 	\label{moti}
% \end{figure}
These methods require training data from both the content and source domains. This issue limits the real-world applications of face stylization due to privacy issues and expensive copyright costs.
\begin{figure*}
	\begin{center}
		\includegraphics[scale=0.75]{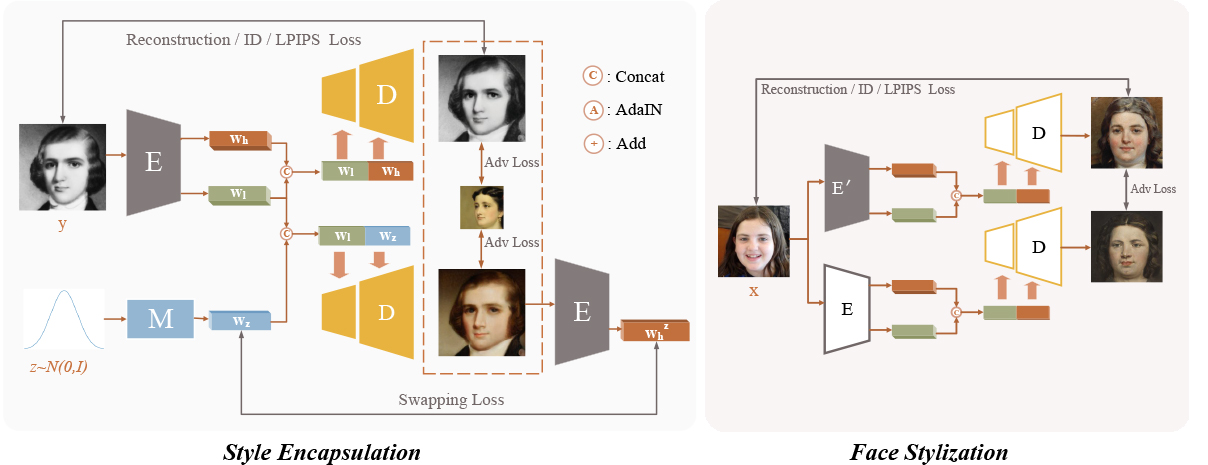} 
	\end{center}
	\caption{Illustration of the style encapsulation training stage. The gray, blue, and yellow parts with notation $\textrm{E}$, $\textrm{M}$, and $\textrm{D}$ denote our encoder, remapping network, and decoder, respectively. 
	The encoder with a standard feature pyramid over a ResNet backbone maps the input image into a latent code with 18 layers. 
	The remapping network with a cascade of several fully connected layers maps the random noise to the higher $\xi$ layers of the latent code.
	The decoder renders a constant vector to an image by the given latent code.}
	\label{net-s1}
\end{figure*}

Inspired by existing privacy-preserving methods\cite{fed1,fed2,fed3,dp1,dpGAN,dp-cgan,pategan,dpboosting,SHOT,liang2021source}, we present MODel-drIven Face stYlization (MODIFY), delivering the style information by a pretrained model to build a bridge between the private source and target dataset.
Generally, the training process of our method consists of two stages, \textit{i.e.}, style encapsulation and face stylization.
At the encapsulation stage, we train an auto-encoder to reconstruct the input with the provided target dataset only.
Specifically, the generative model consists of an FPN encoder~\cite{FPN,psp}, encoding the input images into the latent space;
a remapping network, mapping a known distribution such as Gaussian into the higher-resolution parts of latent space;
and a StyleGAN decoder~\cite{stylegan,stylegan2}, rendering a constant vector to an image with source content and target style by the given latent code.
We enforce the decoder to reconstruct the input from the full code output by the encoder and generate a multimodal version of the input from the fused code, which is a concatenation of the lower-resolution part of the encoder and the remapping network's output.
To this end, we introduce the swapping loss to ensure these two modules have a shared embedding space.
% Intuitively, if we fed the image generated by the fused code into the encoder, the latent code should be a reconstruction of the fused code.
At the stylization stage, with only the source data available, our goal is to learn a mapping from the source data to the latent space without style images.
In the first stage, the encoder has learned a mapping that serves the target data.
To overcome the loss of the identity information caused by the gap between the source and target dataset, we fine-tune the encoder with the adversarial loss, pixel-level reconstruction loss, ID loss~\cite{arcface}, and LPIPS loss~\cite{lpips} while freezing other modules. 
% As the gap is relatively small, this stage would not consume many iterations.
Therefore, MODIFY allows versatile training modes, including online, offline, and test-time training. 
% For example, for the test-time training mode, MODIFY updates the encoder adaptively to capture the content information of the given single input.

To summarize, our contributions are three-fold. 
1) To address the copyright and the privacy limitation in the unsupervised face stylization problem, we first study an interesting and challenging problem—model-driven faced stylization without style images, which avoids the presence of target datasets. 2) We propose a new method called MODIFY, which consists of a generative model and a swapping loss, to store the style information into the proposed model. 
3) We perform quantitative and qualitative evaluations to demonstrate the superiority of MODIFY and its generalizability across different training modes and style domains. 

\section{Method}
The framework, as illustrated in Fig.~\ref{net-s1}, consists of a feature pyramid encoder $\textrm{E}$~\cite{FPN}, a StyleGAN decoder $\textrm{D}$~\cite{stylegan2}, a remapping network $\textrm{M}$ with a cascade of multiple fully-connected layers, and a discriminator $\textrm{Dis}$.
Given an input $\mathbf{x} \in \mathcal{X}$ where $\mathcal{X}$ denotes the source domain, our goal is to translate $\mathbf{x}$ to $\mathbf{x'}\in \mathcal{Y}$ where $\mathcal{Y}$ denotes the target domain.
% MODIFY addresses the problem above through two stages, \textit{i.e.}, style encapsulation with target dataset and face stylization with source dataset, respectively.
% Next, we elaborate our work from these two aspects.

\subsection{Style Encapsulation}

At this stage, only the target (style) dataset $\mathcal{Y}$ is accessible.
As shown in Fig.~\ref{net-s1}, given the input $\mathbf{y}\in \mathcal{Y}$, the encoder outputs a latent code $\mathbf{w}\in \mathcal{W}^{18\times512}$, each $1\times 512$ vector corresponds to one of the 18 AdaIN~\cite{adain} modules of the StyleGAN decoder~\cite{stylegan,stylegan2,Shen2020cvpr,psp}. 
$\mathbf{w}=\{\mathbf{w}_l,\mathbf{w}_h\}$ is obtained by concatenating $\mathbf{w}_l$ and $\mathbf{w}_h$, where $\mathbf{w}_l^{(18-\xi)\times 512}$ is the lower-resolution part, and $\mathbf{w}_h^{\xi\times 512}$ is the higher-resolution part, and $\xi$ is a given hyper-parameter.
Benefiting from the superior disentanglement capability of StyleGAN, $\mathbf{w}_l$ and $\mathbf{w}_h$ describes the output's content and style, respectively.
In this manner, the space $\mathcal{W}$ can be disentangled into the content space $\mathcal{W}_c^{(18-\xi)\times 512}$ and style space $\mathcal{W}_s^{\xi\times 512}$, $\mathbf{w}_l$ denotes the content code, and $\mathbf{w}_h$ denotes the style code.
We try to reconstruct the input strictly, enforcing the decoder to map the $\mathcal{W}$ space into the target data domain, and the encoder to map the input into the $\mathcal{W}$ space. 
During the process of reconstruction, inspired by some theoretical works \cite{he2009robust, he2010principal}, we follow the setting of PSP~\cite{psp}, adding the reconstruction loss $\mathcal{L}_\textrm{r}$ to ensure the pixel-level correspondence, the LPIPS loss $\mathcal{L}_\textrm{lp}$ ~\cite{lpips} to keep the content information, and the ID loss $\mathcal{L}_\textrm{id}$~\cite{arcface} to preserve the identity of the input:
\begin{equation}\label{recon1}
\begin{aligned}
\mathcal{L}_\textrm{r}&=\|\mathbf{y}-\mathbf{y}_r\|_2,\\ 
% \end{equation}
% \begin{equation} \label{lpips1}
\mathcal{L}_{\textrm{lp}}&=\|\textrm{F}(\mathbf{y})-\textrm{F}(\mathbf{y}_r)\|_2,\\
% \end{equation}
% \begin{equation} \label{id1}
\mathcal{L}_\textrm{id}&=1-\langle \textrm{R}(\mathbf{y}),\textrm{R}(\mathbf{y}_r)\rangle,
\end{aligned}
\end{equation}
where $\mathbf{y}_r=\textrm{D}(\textrm{E}(\mathbf{y}))$ is the reconstruction of $\mathbf{y}$, $\textrm{F}(\cdot)$ denotes the perceptual feature extractor, $\langle \mathbf{a},\mathbf{b}\rangle$ denotes the cosine distance between $\mathbf{a}$ and $\mathbf{b}$, and $\textrm{R}(\cdot)$ denotes the pretrained ArcFace~\cite{arcface} network for face recognition.

Additionally, the adversarial loss helps reduce the blur and artifacts, and we employ WGAN~\cite{WGAN} here:
\begin{equation}\label{adv1}
% \mathcal{L}_{\adv}(\hat{y},y_r)=E[log(D(\hat{y}))]+E[log(1-D(y_r)))],
% \min_{\textrm{E},\textrm{D},\textrm{M}}\max_{\textrm{Dis}}
\mathcal{L}_{\textrm{adv}}^{\textrm{r}} = \mathbb{E}[\textrm{Dis}(\mathbf{y'})]-\mathbb{E}[\textrm{Dis}(\mathbf{y}_r)],
\end{equation}
where $\mathbf{y'}\in\mathcal{Y}$ is a re-sampled target image, which is different from the input image $\mathbf{y}$.
% where $\mathbf{\hat{y}}$ is another random-selected target image different from $\mathbf{y}$, $\theta$ denotes the parameters of the generative model, and $\mathbf{c}$ is the clipping parameter.

% \textbf{Swapping Loss} 

We use the remapping network $\textrm{M}$ to capture the distribution of the higher-resolution style code. Given $z \in \mathcal{N}$, where $\mathcal{N}$ is the standard Gaussian distribution, $\textrm{M}$ maps $z$ into $\mathbf{w}_z=\textrm{M}(\mathbf{z})$. Then, we get the fused code $\mathbf{w}_{\textrm{fused}}=\{\mathbf{w}_l,\mathbf{w}_z\}$ and we feed it into the decoder:
\begin{equation}
    \mathbf{y}_z=\textrm{D}(\mathbf{w}_{\textrm{fused}}),
\end{equation}
where $\mathbf{y}_z$ and $\mathbf{y}_r$ have the same content code and different style code.
Thus $\mathbf{y}_z$ is a multimodal version of $\mathbf{y}_r$. 

Inspired by Nie \textit{et al.}~\cite{semi-stylegan}, to enforce that the higher-resolution parts of $\mathcal{W}$ and the embedding space of the remapping network $\textrm{M}$ share a common distribution, we introduce the swapping loss:
\begin{equation} \label{CRloss}
\mathcal{L}_\textrm{swap}=\|\mathbf{w}_z-\mathbf{w}'_z\|_2,
\end{equation}
where $\mathbf{w}'_{\textrm{fused}}=\{\mathbf{w}'_l,\mathbf{w}'_z\}=\textrm{E}(\mathbf{y}_z)$.
The swapping loss ensure that the remapping network learns a meaningful mapping, and avoid mode collapse of the $\textrm{M}$.

The adversarial loss is also imposed on $\mathbf{y}_z$:
\begin{equation}\label{adv1yz}
% \mathcal{L}_{\adv}(\hat{y},y_z)=E[log(D(\hat{y}))]+E[log(1-D(y_z)))],
% \min_{\textrm{E},\textrm{D},\textrm{M}}\max_{\textrm{Dis}}
\mathcal{L}_{\textrm{adv}}^{\textrm{z}}=\mathbb{E}[\textrm{Dis}(\mathbf{y}')]-\mathbb{E}[\textrm{Dis}(\mathbf{y}_z)].
\end{equation}
% where $\mathbf{y}'\in \mathcal{Y}$ is a re-sampled target image.

% \textbf{Weights-adjustment Training}
In summary, the full objective function at this stage can be defined as:
\begin{equation} 
\begin{aligned}
\min_{\textrm{E},\textrm{D},\textrm{M}}\max_{\textrm{Dis}}\mathcal{L}_{\textrm{s1}}=&
\lambda _{\textrm{adv}}^{\textrm{r}}\mathcal{L}_{\textrm{adv}}^{\textrm{r}} +\lambda  _{\textrm{adv}}^{\textrm{z}}\mathcal{L}_{\textrm{adv}}^{\textrm{z}} \\&
+ \lambda _\textrm{r}\mathcal{L}_\textrm{r} + \lambda_\textrm{lp}\mathcal{L}_\textrm{lp} + 
\lambda_\textrm{id}\mathcal{L}_\textrm{id} \\&
+\lambda_\textrm{swap}\mathcal{L}_\textrm{swap},
\end{aligned}
\end{equation}
where $\lambda_{\textrm{adv}}^{\textrm{r}},\lambda_{\textrm{adv}}^{\textrm{z}},\lambda_\textrm{r},\lambda_\textrm{lp},\lambda_\textrm{id},\lambda_\textrm{swap}$ are weighting parameters of different loss functions.

\begin{figure*}[!htbp]
	\footnotesize
	\centering
	\setlength\tabcolsep{0mm}{
	\renewcommand\arraystretch{1}
    \begin{tabular}{p{2cm}<{\centering}p{1.8cm}<{\centering}p{1.8cm}<{\centering}p{1.8cm}<{\centering}p{1.8cm}<{\centering}p{1.8cm}<{\centering}p{1.8cm}<{\centering}p{1.8cm}<{\centering}p{1.8cm}<{\centering}}
    % \hline
    (a) Input & (b) OST & (c) StarGANv2 & (d) StarGANv2 &(e) PSP* & (f) PSP* & (g) MODIFY & (h) MODIFY & (i) MODIFY \\
	\hline
	\# Samples in $\mathcal{X}$ & 1 & 1 & All & 1 & All & 1 & All & All\\
	\hline
	Data-driven &  True & True & True & True & True & False & False & False\\
	\hline
	Training mode & Offline & Offline & Offline & Offline & Offline & Test-time & Online & Offline\\
	\hline
	\includegraphics[width=1.8cm]{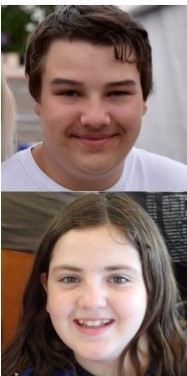} &
	\includegraphics[width=1\linewidth]{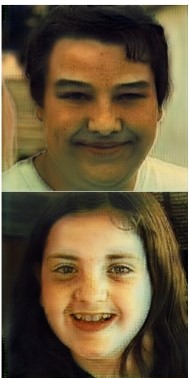} &
	\includegraphics[width=1\linewidth]{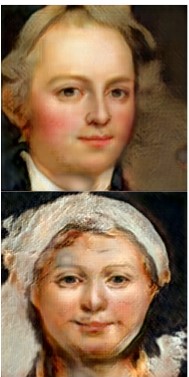} &
	\includegraphics[width=1\linewidth]{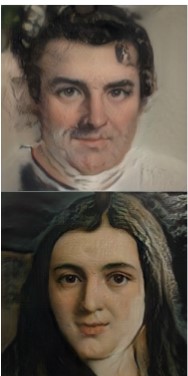} &
	\includegraphics[width=1\linewidth]{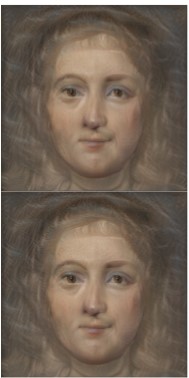} &
	\includegraphics[width=1\linewidth]{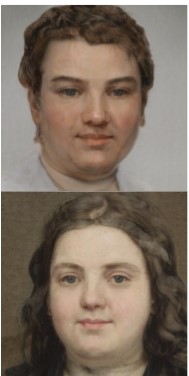} &
	\includegraphics[width=1\linewidth]{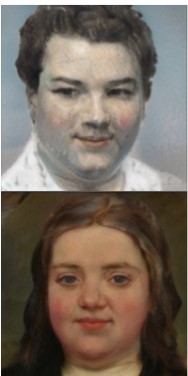} &
	\includegraphics[width=1\linewidth]{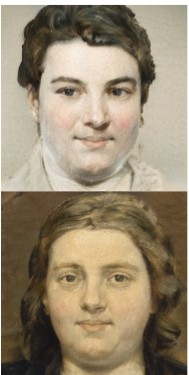} &
	\includegraphics[width=1\linewidth]{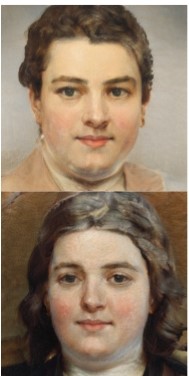} \\
% 	\hline
	\end{tabular}}
	\caption{Results of online, offline and test-time training version of MODIFY compared to other three state-of-the-art approaches (\textit{i.e.}, OST~\cite{OST}, StarGAN v2~\cite{starganv2}, and PSP~\cite{psp}, PSP* notates the unpaired version of PSP).} 
	\label{SOTA}
\end{figure*}

\subsection{Face Stylization}
At this stage, we are only able to access the source dataset.
To avoid identity shifting, we fine-tune the encoder to make it adapt to the unseen source domain.
As illustrated in Fig.~\ref{net-s1}, we replicate the trained encoder, one notated as $\textrm{E}'$ is then fine-tuned to map the input $\mathbf{x}$ into the latent space $\mathcal{W}$, and the other notated as $\textrm{E}$ to be fixed to provide pseudo ground truth in the adversarial loss.
The generative stream is similar to the first stage, the input $\mathbf{x}$ from source dataset is fed to the encoder $\textrm{E}'$ and decoder $\textrm{D}$ in order to obtain the input image with target style $\mathbf{x'}=\textrm{D}(\textrm{E}'(\mathbf{x}))$.
The remapping network does not participate in the training at this stage, and the decoder is frozen.
Besides, our model supports offline, online, and test-time training.

\textbf{Offline Training.}
In the standard training mode, \textit{i.e.}, offline training, the entire training set is available during the training time, and only when training is completed can the model be used for prediction.
We first introduce the objective function of the standard version:
\begin{equation}\label{lossl2}
\begin{aligned}
\min_{\textrm{E}'}\max_{\textrm{Dis}}\mathcal{L}_\textrm{s2}&=
\lambda _\textrm{adv}^\textrm{x}\mathcal{L}_\textrm{adv}^\textrm{x} + \lambda _\textrm{r}\mathcal{L}_\textrm{r} + \lambda_\textrm{lp}\mathcal{L}_\textrm{lp}
+\lambda_\textrm{id}\mathcal{L}_\textrm{id},
\end{aligned}
\end{equation}
where the adversarial loss $\mathcal{L}_\textrm{adv}^\textrm{x}$ is imposed on $\mathbf{x'}$ and the reconstruction loss $\mathcal{L}_\textrm{r}$, LPIPS loss $\mathcal{L}_\textrm{lp}$ and the ID loss $\mathcal{L}_\textrm{id}$ are calculated between $\mathbf{x}$ and $\mathbf{x'}$.

\textbf{Online Training.}
% Therefore, the loss function is calculated on a batch of images, a large batch size is usually more beneficial for the optimizing of networks.
Online training mode process data sequentially. The model is continuously updated during operation as more training data arrives.
Briefly, we only have one image in an iteration.
MODIFY solves the same optimization problem as in Eq. (\ref{lossl2}) with batch size 1.

\textbf{Test-time Training.}
Existing image translation methods remain notoriously weak at generalization under distribution shifts. Even seemingly minor differences between training and test data turn out to defeat state-of-the-art models~\cite{cifar10}.
Test-time training~\cite{test-time} is a new generation mode that does not anticipate the distribution shifts but instead learns from them at test time.
When a test source input arriving, MODIFY fine-tunes the encoder adaptively for this specific input, and the loss function is also the same as Eq. (\ref{lossl2}).
Unlike other training modes, only 50 iterations are required in test-time training, which is acceptable for the test efficiency.

\section{Experiments}
\noindent
\textbf{Implementation details.}
In the style encapsulation stage, we train the model on a public dataset MetFaces~\cite{metface}. There are 1336 face images extracted from works of art in this dataset.
% The implementation details are shown in the supplementary materials.
We adopt a weight decay training method. Specifically, $\{\lambda_{\textrm{swap}},\lambda_{\textrm{lp}},\lambda_{\textrm{adv}}^{\textrm{r}},\lambda_{\textrm{adv}}^{\textrm{z}},\lambda_{\textrm{r}},\lambda_{\textrm{id}}\}=\{0,0.8,0.1,0,0.8,1\}$ during the first 150,000 iterations, while $\{1.0, 0, 0,0.1,0,0\}$ in the next 20,000 iterations.
In the face stylization stage, to balance the samples number of two domains, we select 2000 photos from a high-quality face dataset FFHQ~\cite{stylegan}, and set the weights $\{\lambda_{\textrm{r}}, \lambda_{\textrm{lp}}, \lambda_{\textrm{id}}, \lambda_{\textrm{adv}}^{\textrm{x}}\}=\{0.5, 0.8, 1, 0.01\}$. 
In both stages, we use Adam~\cite{adam} optimizer with $\beta_1=0.9,\beta_2=0.999$, and the learning rate is $1e-4$.
The model is trained for 170,000 steps with a batch size of 4 in the first stage.
In the second stage, the model is trained for 20,000 steps for offline and online training and 50 steps for test-time training. 
We use the MindSpore framework during our implementation, and train our model on RTX TITAN GPU with 24 GB of memory. 
% We show some experiments below, and more results can be seen in our supplementary material.
\begin{table}
	\caption{The quantitative results. 
	The second column is FID (lower is better). 
	%In this paper we calculate FIDs using 2,000 images drawn randomly from the training set. 
	The third column is the voting results of our user study (higher is better).
	}\label{FID}
	\renewcommand\arraystretch{1}
	\center
	\setlength{\tabcolsep}{1.5mm}{
		\begin{tabular}{c|c|c}
			\hline
			Method & FID $\downarrow$ & Percentage $\uparrow$\\
			\hline
			StarGAN v2~\cite{starganv2}&65.5&8.9\%\\
			Unparied PSP~\cite{psp}&74.2&45.9\%\\
			MODIFY&58.6&45.2\%\\
			\hline
		\end{tabular}
	}
\end{table}

\subsection{Comparisons with State-of-the-art}
Since this is the first model-driven translation method, 
we evaluate three existing state-of-the-art methods under a data-driven condition \textbf{except ours} for comparison.
We compare MODIFY, trained in online, offline, and test-time modes, 
to the one-shot translation method OST~\cite{OST}, the one-shot and standard versions of StarGAN v2~\cite{starganv2} and PSP~\cite{psp}.
The one-shot versions of these two methods are trained on one source image and the whole target dataset.
Note that our datasets are unpaired, which does not match the setting of PSP.
Therefore, we modify the ID loss, L2 loss, and LPIPS loss of PSP to be calculated between input and output instead of between output and the ground truth in the original version. 
We also add a discriminator to calculate the adversarial loss between the output and the target images, thereby improving PSP to an unpaired version.
\textbf{The unpaired PSP could be seen as a data-driven version of our MODIFY, which can be considered as the best effect that we can achieve}.

\noindent
\textbf{Qualitative Results.}
Fig.~\ref{SOTA} shows the qualitative comparison results.
Fig.~\ref{SOTA} (d), (f), (h) and (i) use the entire source dataset, and Fig.~\ref{SOTA} (b), (c), (e) and (g) use one sample in source dataset.
Note that Fig.~\ref{SOTA} (b), (c), (e) use a random-selected source image for training, and our method shown in Fig.~\ref{SOTA} (i) adaptively updates the model for the input test image without separate training.
In the one-shot scenario, OST preserves too much input information, thereby ignoring the style transfer. Extremely little data leads to mode collapse in unpaired PSP. StarGAN v2 still performs very well in style transfer but preserves poorly the identity of input.
These three methods can only learn a single seen source image during training, therefore they cannot perfectly predict all unseen test images.
In comparison, our test-time training mode is more flexible therefore more robust.
% In the standard data-driven version, both StarGAN v2 and unpaired PSP perform well on identity preserving and style rendering. 
% And each training mode of MODIFY achieves a similar effect, even with harder conditions.
% \begin{figure}
% 	\begin{center}
% 		\includegraphics[scale=0.31]{results/ablation.pdf} 
% 	\end{center}
% 	\caption{Ablation study of the swapping loss.}
% 	\label{ablation}
% \end{figure}

\begin{figure}
    % \tiny
	\centering
	\setlength\tabcolsep{0.1mm}{
	\renewcommand\arraystretch{1.2}
    \begin{tabular}{p{1.5cm}<{\centering}p{3cm}<{\centering}p{3cm}<{\centering}}
% 	\hline
	(a) Input & (b) without $\mathcal{L}_{\textrm{swap}}$ &(c) with $\mathcal{L}_{\textrm{swap}}$\\
	\includegraphics[width=1\linewidth]{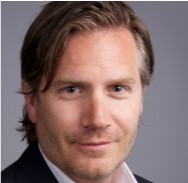} &
	\includegraphics[width=1\linewidth]{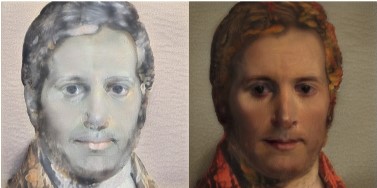} &
	\includegraphics[width=1\linewidth]{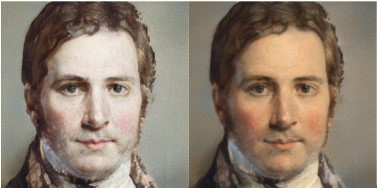} \\
% 	\hline
	\end{tabular}}
	\caption{Ablation study of the swapping loss $\mathcal{L}_{\textrm{swap}}$.} 
	\label{ablation}
\end{figure}

% \begin{figure}
% 	\begin{center}
% 		\includegraphics[scale=0.26]{results/ablation-L.pdf} 
% 	\end{center}
% 	\caption{Ablation study of the fusion layer $\xi$.}
% 	\label{ablation-L}
% \end{figure}
\begin{figure}
    \small
	\centering
	\setlength\tabcolsep{0mm}{
	\renewcommand\arraystretch{1.2}
    \begin{tabular}{p{1.4cm}<{\centering}p{1.4cm}<{\centering}p{1.4cm}<{\centering}p{1.4cm}<{\centering}p{1.4cm}<{\centering}p{1.4cm}<{\centering}}
% 	\hline
	(a) Input & (b) $\xi=3$ & (c) $\xi=6$ & (d) $\xi=9$ & (e) $\xi=12$ & (f) $\xi=15$\\
	\includegraphics[width=1\linewidth]{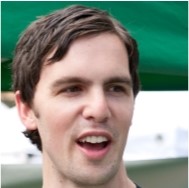} &
	\includegraphics[width=1\linewidth]{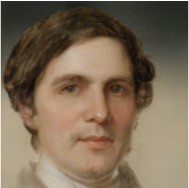} &
	\includegraphics[width=1\linewidth]{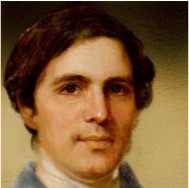} &
	\includegraphics[width=1\linewidth]{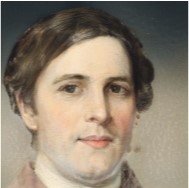} &
	\includegraphics[width=1\linewidth]{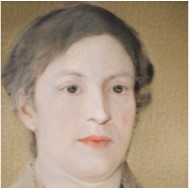} &
	\includegraphics[width=1\linewidth]{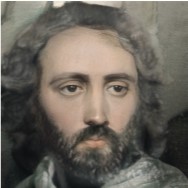} \\
% 	\hline
	\end{tabular}}
	\caption{Ablation study of the fusion layer $\xi$.} 
	\label{ablation-L}
\end{figure} 

% \begin{figure}
% 	\begin{center}
% 		\includegraphics[scale=0.12]{results/s1-results.pdf} 
% 	\end{center}
% 	\caption{Ablation study on the second-stage training.}
% 	\label{s1-results}
% \end{figure}
% \begin{figure}
%     \small
% 	\centering
% 	\setlength\tabcolsep{0.1mm}{
% 	\renewcommand\arraystretch{1.2}
%     \begin{tabular}{p{2.2cm}<{\centering}p{2.5cm}<{\centering}p{2.2cm}<{\centering}}
% % 	\hline
% 	(a) Input & (b) without Stage 2 &(c) with Stage 2\\
% 	\includegraphics[width=1\linewidth]{results/abs-a.jpg} &
% 	\includegraphics[width=2.2cm]{results/abs-b.jpg} &
% 	\includegraphics[width=1\linewidth]{results/abs-c.jpg} \\
% % 	\hline
% 	\end{tabular}}
% 	\caption{Ablation study on training the second stage.} 
% 	\label{s1-results}
% \end{figure}

\noindent
\textbf{Quantitative Results.}
We conduct quantitative experiments using two widely-used metrics, Fréchet Inception Distance (FID) and user study.
We compare the FIDs on the standard version of StarGAN v2 and unpaired PSP to the offline version of our method.
As shown in the second column of Table~\ref{FID}, we achieve the best results even under a more brutal model-driven condition.
The poor performance of unpaired PSP is not because of poor image quality, but the lack of diversity, thereby cannot capture the entire target distribution.
Then we investigate user studies against the state-of-the-art methods.
Ten random-selected test photos and corresponding outputs synthesized by these three methods are presented to 100 subjects, who are told to vote their favorite one.
The results are shown in the third column of Table~\ref{FID}.
The unpaired PSP gets a 45.9\% approval rate, and we get 45.2\%.
As we mentioned before, the unpaired PSP is a data-driven version of our method, implying the best quality that we can achieve in theory.

% \begin{figure}
% 	\begin{center}
% 		\includegraphics[scale=0.13]{results/interpolate.pdf} 
% 	\end{center}
% 	\caption{}
% 	\label{interpolate}
% \end{figure}
% \begin{figure}
%     \tiny
% 	\centering
% 	\setlength\tabcolsep{0mm}{
% 	\renewcommand\arraystretch{1.2}
%     \begin{tabular}{p{1.2cm}<{\centering}p{1.2cm}<{\centering}p{1.2cm}<{\centering}p{1.2cm}<{\centering}p{1.2cm}<{\centering}p{1.2cm}<{\centering}p{1.2cm}<{\centering}}
% % 	\hline
% 	(a) Input & (b) Style 1 & (c) Style 2 & (d) Style 3 & (e) Style 4 & (f) Style 5 & (g) Style 6\\
% 	\includegraphics[width=1\linewidth]{results/inter-a.jpg} &
% 	\includegraphics[width=1\linewidth]{results/inter-b.jpg} &
% 	\includegraphics[width=1\linewidth]{results/inter-c.jpg} &
% 	\includegraphics[width=1\linewidth]{results/inter-d.jpg} &
% 	\includegraphics[width=1\linewidth]{results/inter-e.jpg} &
% 	\includegraphics[width=1\linewidth]{results/inter-f.jpg} &
% 	\includegraphics[width=1\linewidth]{results/inter-g.jpg} \\
% % 	\hline
% 	\end{tabular}}
% 	\caption{Interpolation results between two randomly-sampled noise vectors.} 
% 	\label{interpolate}
% \end{figure} 

\subsection{Ablation Study}
% We analyze the effects of swapping loss and the fusion layer in the proposed model.
% We then prove the importance of the second-stage training on identity preserving.

\noindent
\textbf{Swapping Loss.}
To validate the importance of the swapping loss, we train a variant of MODIFY for comparison by removing $\mathcal{L}_{\textrm{swap}}$, and show the results in Fig.~\ref{ablation}. Without the swapping loss, the embedding spaces of the remapping network and encoder are different, resulting in the style information not being correctly encapsulated in the remapping network.
Thereby there are too many artifacts in the generated images.

\noindent
\textbf{Fusion Layer.}
The hyper-parameter $\xi$ indicates which layer the style code output by the remapping network $\textrm{M}$ is fused with the latent code output by the encoder $\textrm{E}$.
To explore the influence of $\xi$ on the synthesized image, we train several versions of MODIFY with different $\xi$.
As shown in Fig.~\ref{ablation-L}, the larger the $\xi$, the more the identity is lost.

\section{Conclusion}
In this paper, we propose MODIFY with two training stages, to solve the face stylization problem under a data privacy condition.
In the first stage, MODIFY trains a generative model with an FPN encoder, a StyleGAN decoder, and a remapping network to disentangle and preserving the style information.
We propose a swapping loss to enforce that the encoder and the remapping network share a common embedding space.
In the second stage, MODIFY adapts the model to the unseen source domain.
Our method is extremely simple and satisfies versatile training modes for the face stylization stage.
% We hope our method
% However, since only the target dataset is accessible in the style encapsulation stage, MODIFY encounters troubles of processing parts that are only present in the source domain, such as the glasses, tattoo, and exaggerated emotion in the MetFaces dataset.
% That is a challenging issue that remains as our future work to explore.
\section{Acknowledgements}
This work is funded by National Natural Science Foundation of China (Grant No. 62276256, 62206277) and CAAI-Huawei MindSpore Open Fund.

% -------------------------------------------------------------------------
\small
\bibliographystyle{icassp}
\bibliography{icassp}

\begin{thebibliography}{10}

\bibitem{warpgan}
Yichun Shi, Debayan Deb, and Anil~K Jain,
\newblock ``Warpgan: Automatic caricature generation,''
\newblock in {\em Proc. CVPR}, 2019.

\bibitem{psp}
Richardson Elad, Alaluf Yuval, Patashnik Or, Nitzan Yotam, Azar Yaniv, Shapiro
  Stav, and Cohen-Or Daniel,
\newblock ``Encoding in style: a stylegan encoder for image-to-image
  translation,''
\newblock in {\em Proc. CVPR}, 2021.

\bibitem{Men_2022_CVPR}
Yifang Men, Yuan Yao, Miaomiao Cui, Zhouhui Lian, Xuansong Xie, and Xian-Sheng
  Hua,
\newblock ``Unpaired cartoon image synthesis via gated cycle mapping,''
\newblock in {\em Proc. CVPR}, 2022.

\bibitem{cyclegan}
Jun-Yan Zhu, Taesung Park, Phillip Isola, and Alexei~A Efros,
\newblock ``Unpaired image-to-image translation using cycle-consistent
  adversarial networks,''
\newblock in {\em Proc. ICCV}, 2017.

\bibitem{Li_2022_CVPR}
Zhiheng Li, Martin~Renqiang Min, Kai Li, and Chenliang Xu,
\newblock ``Stylet2i: Toward compositional and high-fidelity text-to-image
  synthesis,''
\newblock in {\em Proc. CVPR}, 2022.

\bibitem{fed1}
Jakub Kone{\v{c}}n{\`y}, H~Brendan McMahan, Daniel Ramage, and Peter
  Richt{\'a}rik,
\newblock ``Federated optimization: Distributed machine learning for on-device
  intelligence,''
\newblock in {\em arXiv preprint arXiv:1610.02527}, 2016.

\bibitem{fed2}
Jakub Kone{\v{c}}n{\`y}, H~Brendan McMahan, Felix~X Yu, Peter Richt{\'a}rik,
  Ananda~Theertha Suresh, and Dave Bacon,
\newblock ``Federated learning: Strategies for improving communication
  efficiency,''
\newblock in {\em arXiv preprint arXiv:1610.05492}, 2016.

\bibitem{fed3}
H~Brendan McMahan, Eider Moore, Daniel Ramage, and Blaise~Ag{\"u}era y~Arcas,
\newblock ``Federated learning of deep networks using model averaging,''
\newblock in {\em arXiv preprint arXiv:1602.05629}, 2016.

\bibitem{dp1}
Cynthia Dwork, Krishnaram Kenthapadi, Frank McSherry, Ilya Mironov, and Moni
  Naor,
\newblock ``Our data, ourselves: Privacy via distributed noise generation,''
\newblock in {\em Annual International Conference on the Theory and
  Applications of Cryptographic Techniques}, 2006.

\bibitem{dpGAN}
Liyang Xie, Kaixiang Lin, Shu Wang, Fei Wang, and Jiayu Zhou,
\newblock ``Differentially private generative adversarial network,''
\newblock in {\em arXiv preprint arXiv:1802.06739}, 2018.

\bibitem{dp-cgan}
Reihaneh Torkzadehmahani, Peter Kairouz, and Benedict Paten,
\newblock ``Dp-cgan: Differentially private synthetic data and label
  generation,''
\newblock in {\em Proc. CVPR}, 2019.

\bibitem{pategan}
Jinsung Yoon, James Jordon, and Mihaela van~der Schaar,
\newblock ``{PATE}-{GAN}: Generating synthetic data with differential privacy
  guarantees,''
\newblock in {\em Proc. ICLR}, 2019.

\bibitem{dpboosting}
Marcel Neunhoeffer, Zhiwei~Steven Wu, and Cynthia Dwork,
\newblock ``Private post-gan boosting,''
\newblock in {\em Proc. ICLR}, 2021.

\bibitem{SHOT}
Jian Liang, Dapeng Hu, and Jiashi Feng,
\newblock ``Do we really need to access the source data? source hypothesis
  transfer for unsupervised domain adaptation,''
\newblock in {\em Proc. ICML}, 2020.

\bibitem{liang2021source}
Jian Liang, Dapeng Hu, Yunbo Wang, Ran He, and Jiashi Feng,
\newblock ``Source data-absent unsupervised domain adaptation through
  hypothesis transfer and labeling transfer,''
\newblock {\em IEEE Transactions on Pattern Analysis and Machine Intelligence},
  vol. 44, no. 11, pp. 8602--8617, 2021.

\bibitem{FPN}
Tsung-Yi Lin, Piotr Doll{\'a}r, Ross Girshick, Kaiming He, Bharath Hariharan,
  and Serge Belongie,
\newblock ``Feature pyramid networks for object detection,''
\newblock in {\em Proc. CVPR}, 2017.

\bibitem{stylegan}
Tero Karras, Samuli Laine, and Timo Aila,
\newblock ``A style-based generator architecture for generative adversarial
  networks,''
\newblock in {\em Proc. CVPR}, 2019.

\bibitem{stylegan2}
Tero Karras, Samuli Laine, Miika Aittala, Janne Hellsten, Jaakko Lehtinen, and
  Timo Aila,
\newblock ``Analyzing and improving the image quality of stylegan,''
\newblock in {\em Proc. CVPR}, 2020.

\bibitem{arcface}
Jiankang Deng, Jia Guo, Niannan Xue, and Stefanos Zafeiriou,
\newblock ``Arcface: Additive angular margin loss for deep face recognition,''
\newblock in {\em Proc. CVPR}, 2019.

\bibitem{lpips}
Richard Zhang, Phillip Isola, Alexei~A Efros, Eli Shechtman, and Oliver Wang,
\newblock ``The unreasonable effectiveness of deep features as a perceptual
  metric,''
\newblock in {\em Proc. CVPR}, 2018.

\bibitem{adain}
Xun Huang and Serge Belongie,
\newblock ``Arbitrary style transfer in real-time with adaptive instance
  normalization,''
\newblock in {\em Proc. ICCV}, 2017.

\bibitem{Shen2020cvpr}
Yujun Shen, Jinjin Gu, Xiaoou Tang, and Bolei Zhou,
\newblock ``Interpreting the latent space of gans for semantic face editing,''
\newblock in {\em Proc. CVPR}, 2020.

\bibitem{he2009robust}
Ran He, Bao-Gang Hu, and Xiao-Tong Yuan,
\newblock ``Robust discriminant analysis based on nonparametric maximum
  entropy,''
\newblock in {\em Proc. ACML}, 2009.

\bibitem{he2010principal}
Ran He, Baogang Hu, XiaoTong Yuan, and Wei-Shi Zheng,
\newblock ``Principal component analysis based on non-parametric maximum
  entropy,''
\newblock {\em Neurocomputing}, vol. 73, no. 10-12, pp. 1840--1852, 2010.

\bibitem{WGAN}
Martin Arjovsky, Soumith Chintala, and L{\'e}on Bottou,
\newblock ``Wasserstein generative adversarial networks,''
\newblock in {\em Proc. ICML}, 2017.

\bibitem{semi-stylegan}
Weili Nie, Tero Karras, Animesh Garg, Shoubhik Debnath, Anjul Patney, Ankit
  Patel, and Animashree Anandkumar,
\newblock ``Semi-supervised stylegan for disentanglement learning,''
\newblock in {\em Proc. ICML}, 2020.

\bibitem{OST}
Sagie Benaim and Lior Wolf,
\newblock ``{One-Shot Unsupervised Cross Domain Translation},''
\newblock in {\em Proc. NeurIPS}, 2018.

\bibitem{starganv2}
Yunjey Choi, Youngjung Uh, Jaejun Yoo, and Jung-Woo Ha,
\newblock ``Stargan v2: Diverse image synthesis for multiple domains,''
\newblock in {\em Proc. CVPR}, 2020.

\bibitem{cifar10}
Benjamin Recht, Rebecca Roelofs, Ludwig Schmidt, and Vaishaal Shankar,
\newblock ``Do cifar-10 classifiers generalize to cifar-10?,''
\newblock in {\em arXiv preprint arXiv:1806.00451}, 2018.

\bibitem{test-time}
Yu~Sun, Xiaolong Wang, Zhuang Liu, John Miller, Alexei Efros, and Moritz Hardt,
\newblock ``Test-time training with self-supervision for generalization under
  distribution shifts,''
\newblock in {\em Proc. ICML}, 2020.

\bibitem{metface}
Tero Karras, Miika Aittala, Janne Hellsten, Samuli Laine, Jaakko Lehtinen, and
  Timo Aila,
\newblock ``Training generative adversarial networks with limited data,''
\newblock in {\em Proc. NeurIPS}, 2020.

\bibitem{adam}
Diederik~P Kingma and Jimmy Ba,
\newblock ``Adam: A method for stochastic optimization,''
\newblock in {\em Proc. ICML}, 2015.

\end{thebibliography}

\end{document}

\sloppy

% Example definitions.
% --------------------
\def\x{{\mathbf x}}
\def\L{{\cal L}}

% Title.
% ------
\title{Model-driven Face Stylization: A privacy-preserving Approach\\
Appendix}
%
% Single address.
% ---------------
\name{Anonymous ICME submission}
%Address and e-mail should NOT be added in the submission paper. They should be present only in the camera ready paper. 
\address{}

\maketitle

\section{Discriminator}
We build our model upon the official Pytorch implementation of PSP~\cite{psp}, from which we inherit most of the training details, including the progressive discriminator proposed by Karras \textit{et al.}~\cite{PG-GAN}.

\begin{figure}
	\begin{center}
		\includegraphics[scale=0.25]{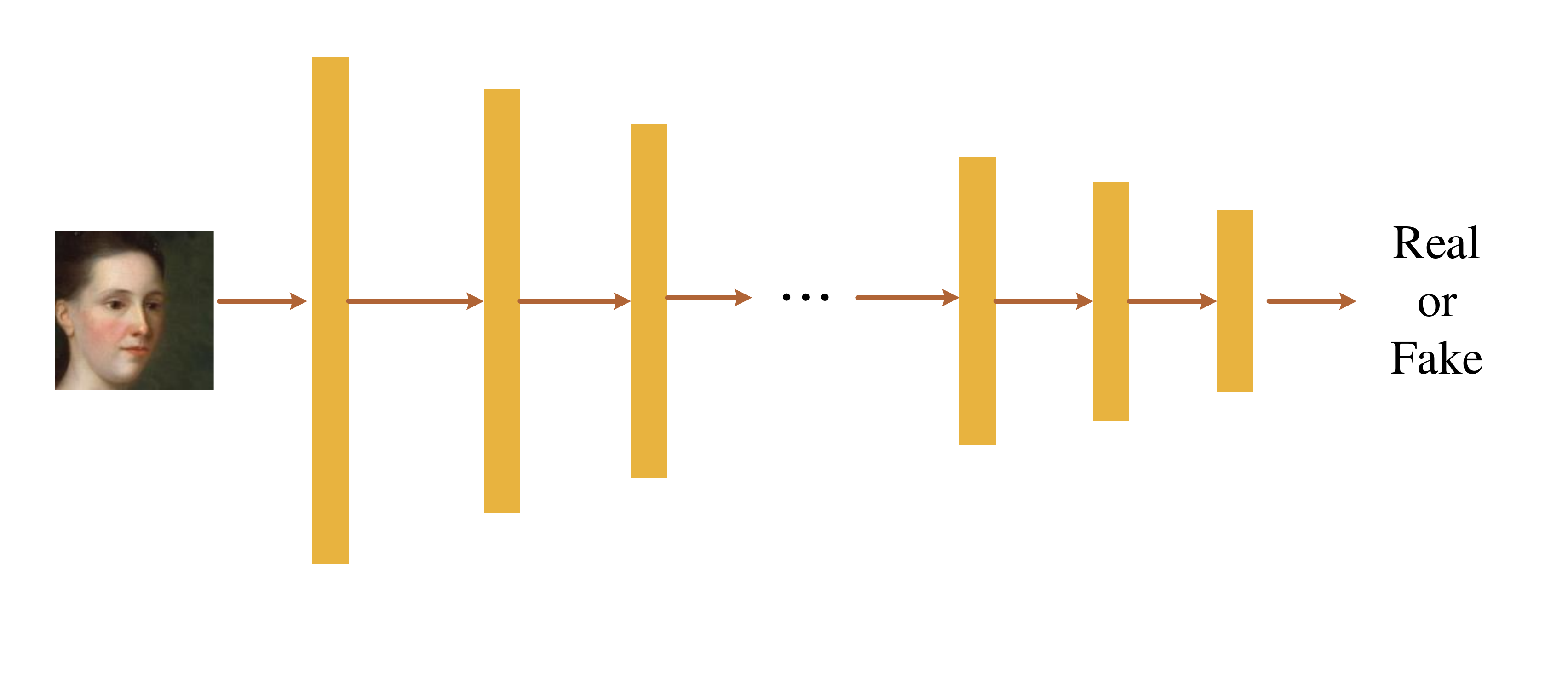} 
	\end{center}
	\caption{The architecture of the discriminator.}
	\label{moti}
\end{figure}

\begin{figure}
	% \small
	\centering
		\includegraphics[width=0.8\linewidth]{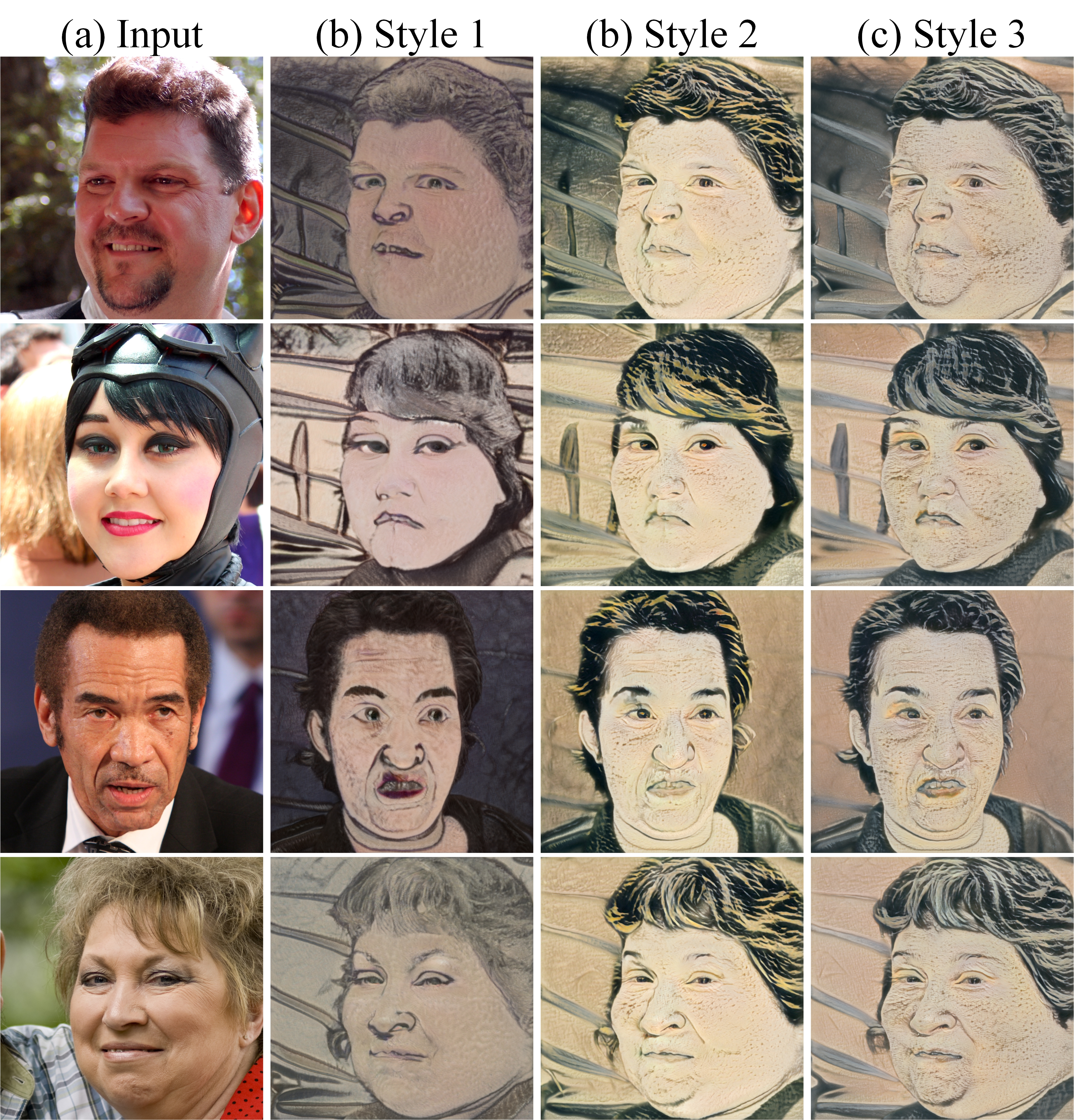} 
	\caption{Results generated from the style-preserving model trained on Ukiyoe~\cite{pinkney2020ukiyoe} dataset with different noises.} 
	\label{uky}
\end{figure}
\section{More Visualized Results}
\begin{figure*}[!htbp]
	\small
	\centering
	\setlength\tabcolsep{0.1mm}{
		\renewcommand\arraystretch{1.5}
		\begin{tabular}{p{2cm}<{\centering}p{1.8cm}<{\centering}p{1.8cm}<{\centering}p{1.8cm}<{\centering}p{1.8cm}<{\centering}p{1.8cm}<{\centering}p{1.8cm}<{\centering}p{1.8cm}<{\centering}p{1.8cm}<{\centering}}
			% \hline
			(a) Input & (b) OST & (c) StarGAN v2 & (d) StarGAN v2 &(e) Unpaired PSP & (f) Unpaired PSP & (g) MODIFY & (h) MODIFY & (i) MODIFY \\
			\hline
			\# Samples in $\mathcal{X}$ & 1 & 1 & All & 1 & All & 1 & All & All\\
			\hline
			Data-driven &  True & True & True & True & True & False & False & False\\
			\hline
			Training mode & Offline & Offline & Offline & Offline & Offline & Test-time & Online & Offline\\
			%			\hline
			\includegraphics[height=0.8\textheight]{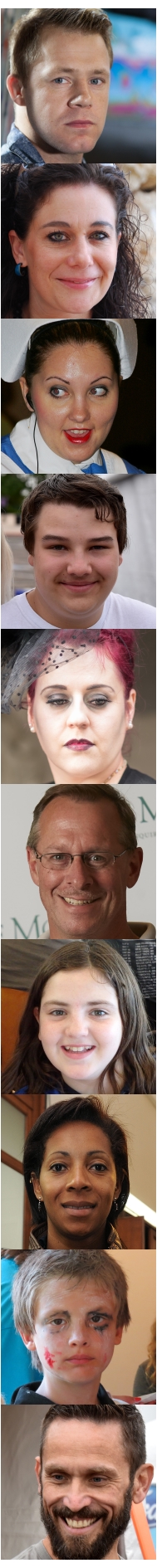} &
			\includegraphics[height=0.8\textheight]{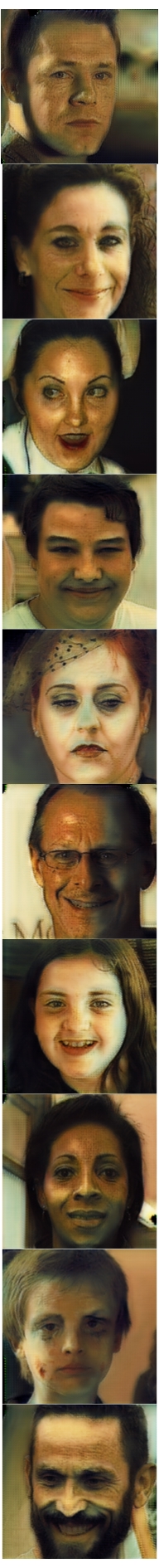} &
			\includegraphics[height=0.8\textheight]{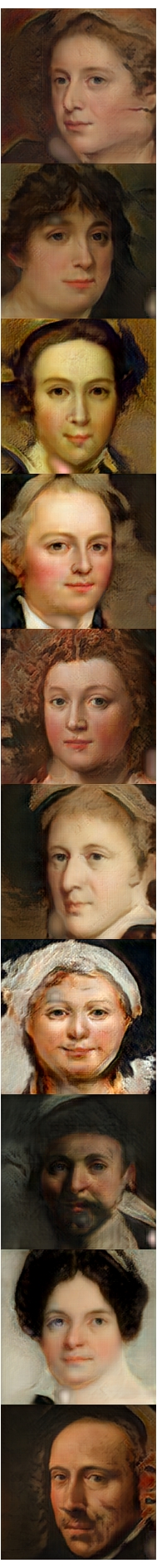} &
			\includegraphics[height=0.8\textheight]{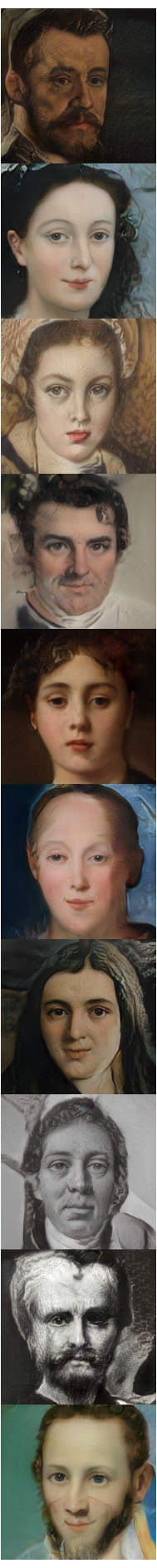} &
			\includegraphics[height=0.8\textheight]{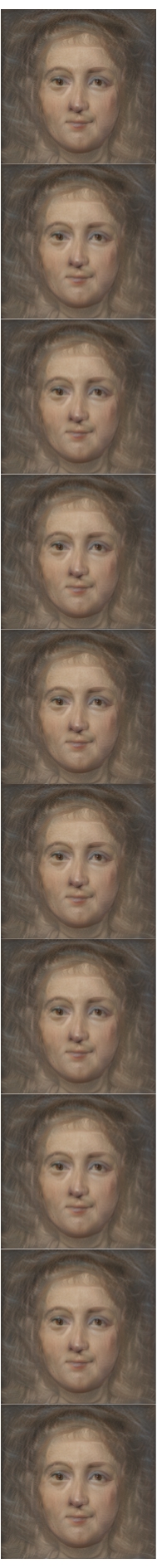} &
			\includegraphics[height=0.8\textheight]{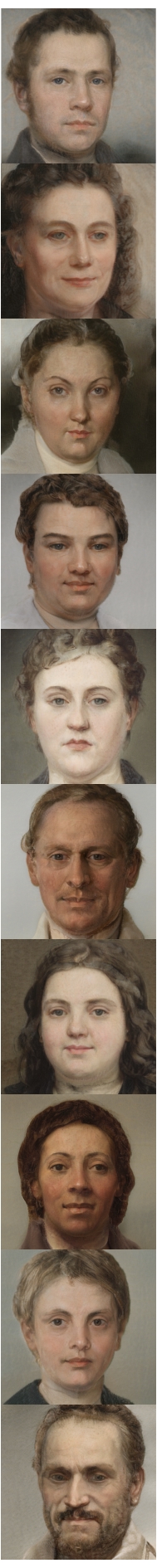} &
			\includegraphics[height=0.8\textheight]{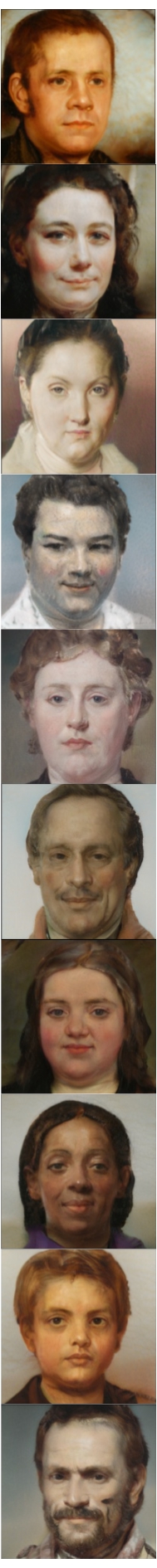} &
			\includegraphics[height=0.8\textheight]{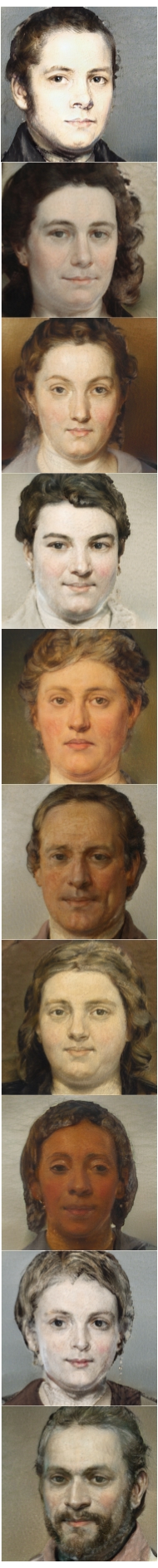} &
			\includegraphics[height=0.8\textheight]{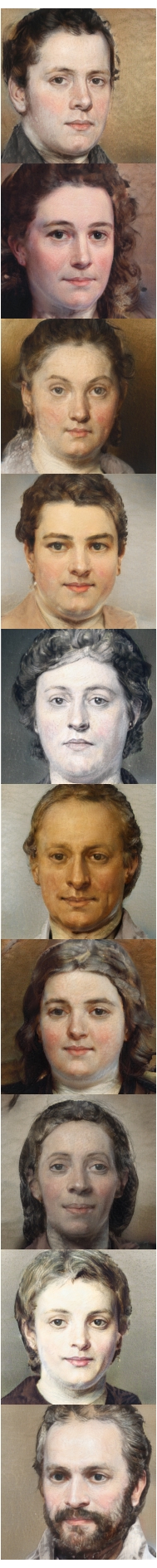} \\
			% 	\hline
	\end{tabular}}
	\caption{Results of online, offline and test-time training versions of MODIFY compared to other three state-of-the-art approaches (\textit{i.e.}, OST~\cite{OST}, StarGAN v2~\cite{starganv2}, and PSP~\cite{psp}). Fig.~\ref{SOTA} (d), (f), (h) and (i) use the entire source dataset, and Fig.~\ref{SOTA} (b), (c), (e) and (g) use one sample in source dataset. Note that  (b), (c), (e) use a random-selected source image for training, and our method shown in (i) adaptively updates the model for the input test image without separate training.} 
	\label{SOTA}
\end{figure*}

\begin{figure*}
\begin{center}
	\includegraphics[scale=0.2]{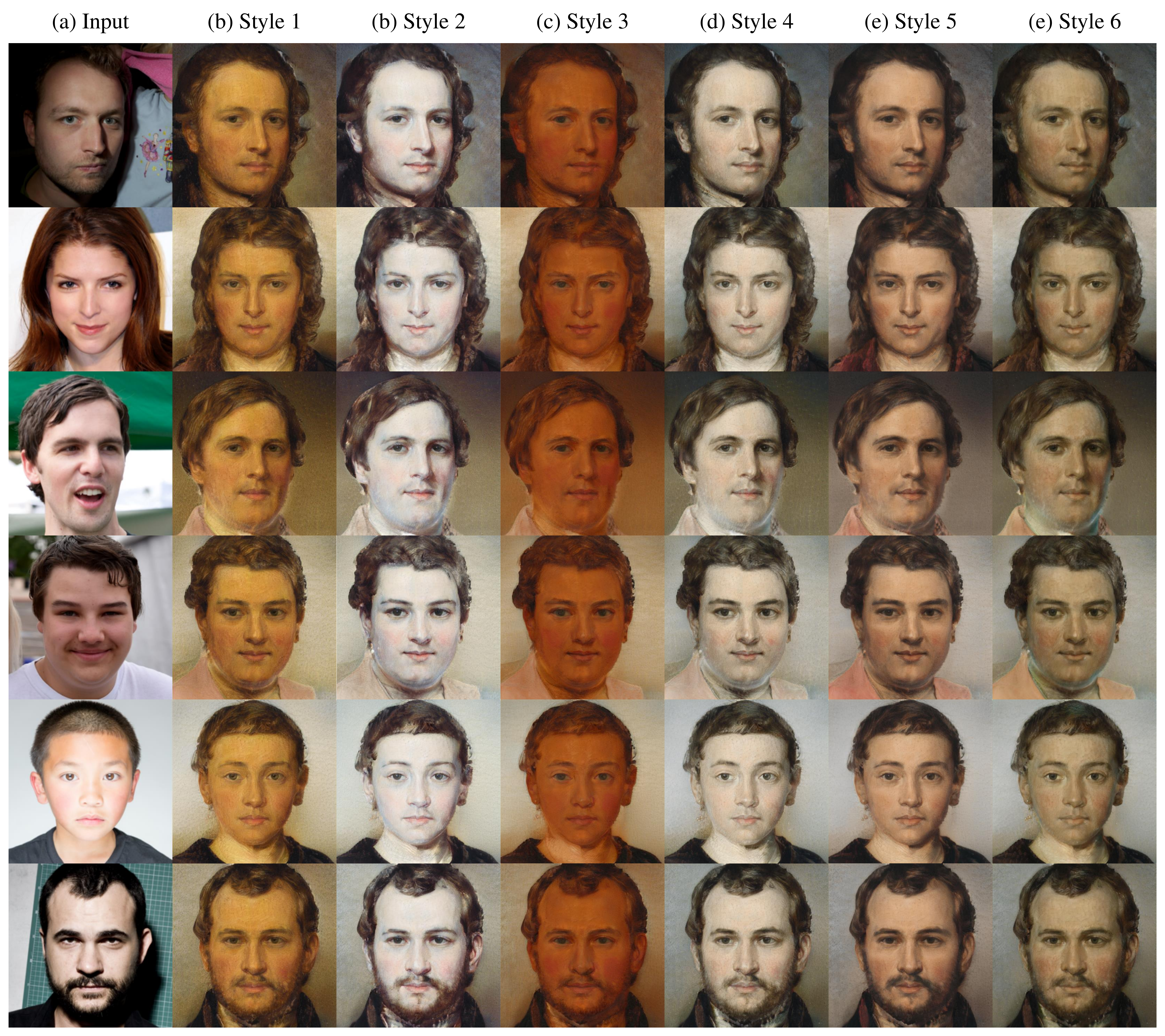} 
\end{center}
\caption{Results generated from different noise vector. Each column generated from the same one. MODIFY can generate multi-style images from different noises.}
\label{rand-noise}
\end{figure*}

\begin{figure}
	% \small
	\centering
	\setlength\tabcolsep{0mm}{
		\renewcommand\arraystretch{1.2}
		\scalebox{0.85}{
			\begin{tabular}{p{2cm}<{\centering}p{2cm}<{\centering}p{2cm}<{\centering}p{2cm}<{\centering}}
				% 	\hline
				(a) Input & (b) Style 1&(c) Style 2& (d) Style 3\\
				\includegraphics[width=1\linewidth]{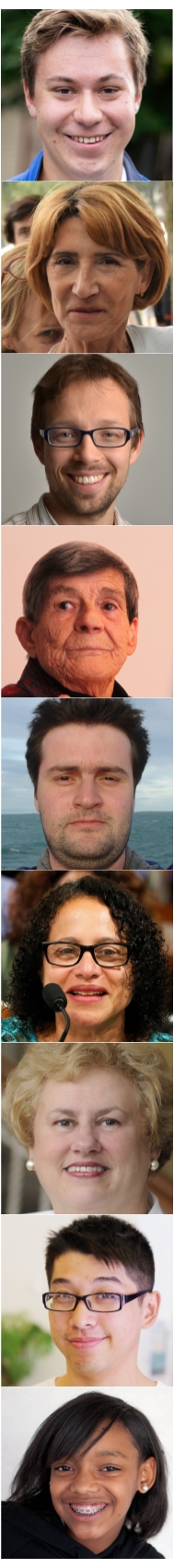} &
				\includegraphics[width=1\linewidth]{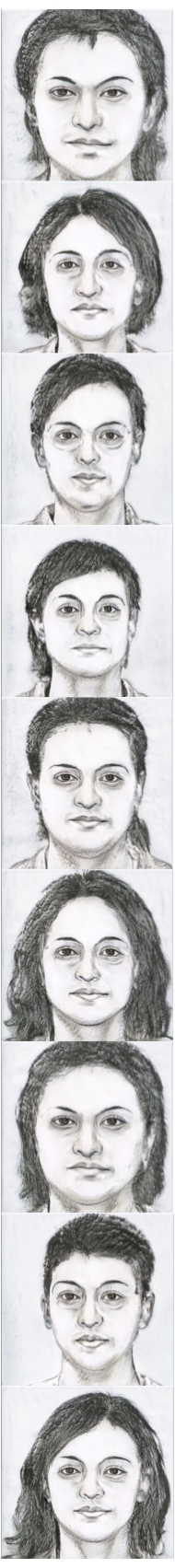} &
				\includegraphics[width=1\linewidth]{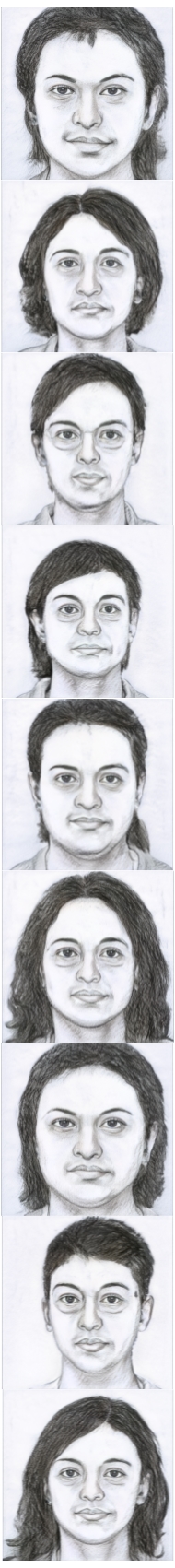} &
				\includegraphics[width=1\linewidth]{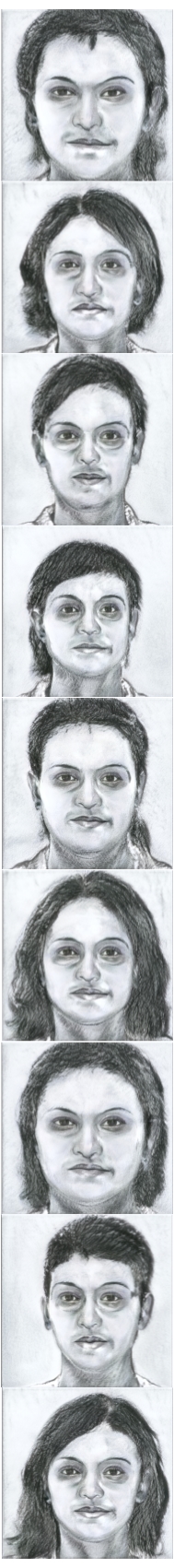} \\
				% 	\hline
	\end{tabular}}}
	\caption{Results generated from the style-preserving model trained on CUFS~\cite{cufs} dataset with different noises.} 
	\label{sketch}
\end{figure}

\begin{figure}
	% \small
	\centering
	\setlength\tabcolsep{0mm}{
		\renewcommand\arraystretch{1.2}
		\scalebox{0.85}{
			\begin{tabular}{p{2cm}<{\centering}p{2cm}<{\centering}p{2cm}<{\centering}p{2cm}<{\centering}}
				% 	\hline
				(a) Input & (b) Style 1&(c) Style 2& (d) Style 3\\
				\includegraphics[width=1\linewidth]{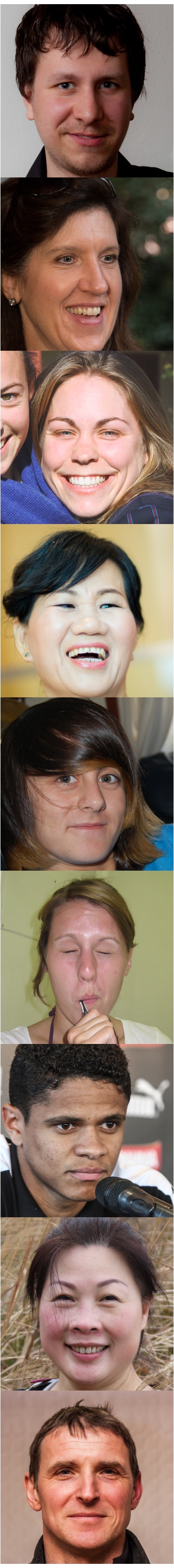} &
				\includegraphics[width=1\linewidth]{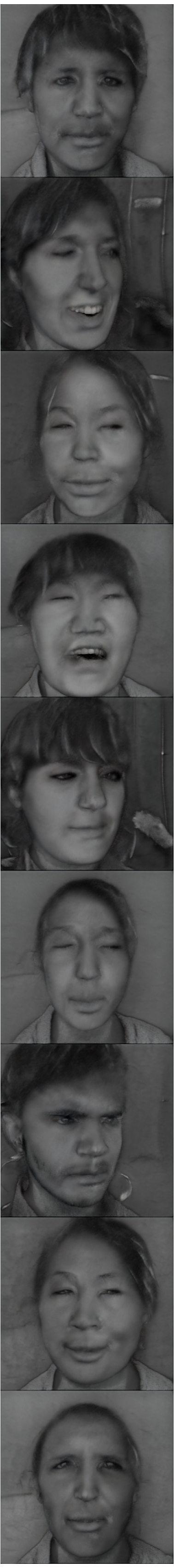} &
				\includegraphics[width=1\linewidth]{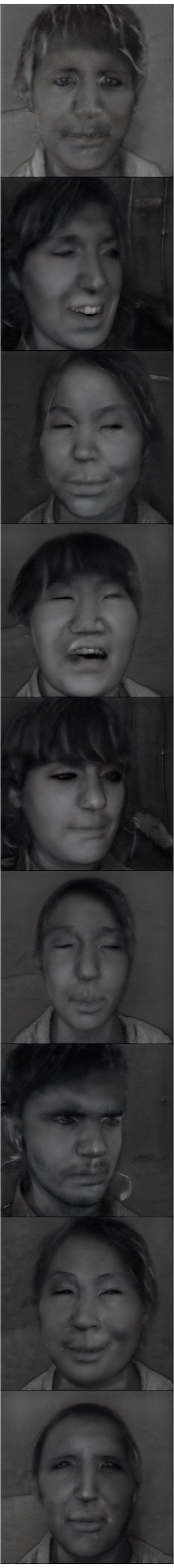} &
				\includegraphics[width=1\linewidth]{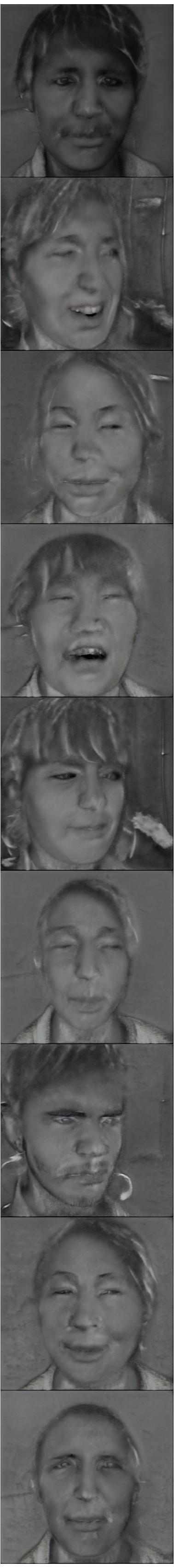} \\
				% 	\hline
	\end{tabular}}}
	\caption{Results generated from the style-preserving model trained on LAMP-HQ~\cite{lamp-hq} dataset with different noises.} 
	\label{NIR}
\end{figure}

We provide more testing results of the comparisons with the state-of-the-art methods in Fig.~\ref{SOTA}, and results generated from the style-preserving model trained on Ukiyoe, CUFS and LAMP-HQ in Fig.~\ref{uky}, Fig.~\ref{sketch} and Fig.~\ref{NIR}, respectively.
Fig.~\ref{rand-noise} shows the noise-guided results of MODIFY, we are able to generate diverse results from random-selected noise vectors.


% -------------------------------------------------------------------------
\bibliographystyle{main}
\bibliography{main}